\documentclass[11pt,a4paper]{article}
\usepackage[hyperref]{acl2019}
\usepackage{times}
\usepackage{latexsym}

\usepackage{url}
\usepackage{verbatim}

\aclfinalcopy % Uncomment this line for the final submission
\newcommand{\todo}[1]{{\color{red}{#1}}}
\usepackage[inline]{enumitem}
\usepackage{amsmath}
\usepackage{amssymb}
\usepackage{bm}
\usepackage{graphicx}
\usepackage{booktabs}
\usepackage{xspace}

\usepackage{subcaption}
\usepackage[skip=0pt]{caption}
\usepackage{xcolor}
\newcounter{todocntjb}

\newcounter{todocnt}

\newcounter{todocntmvdw}

\title{Understanding Multi-Head Attention in Abstractive Summarization}

\author{
\begin{tabular}{c}
Joris Baan$^1$\quad
Maartje ter Hoeve$^2$\quad
Marlies van der Wees$^1$\\[3pt]
Anne Schuth$^1$\quad
Maarten de Rijke$^2$
\end{tabular}
\vspace{3pt}
\\
$^1$DPG Media, Amsterdam\quad
$^2$University of Amsterdam, Amsterdam
\vspace{2pt}
\\
\texttt{\{joris.baan,marlies.van.der.wees,anne.schuth\}@dpgmedia.nl}
\\
\texttt{\{m.a.terhoeve,derijke\}@uva.nl}
}

\date{}

\begin{document}
\maketitle
\begin{abstract}
Attention mechanisms in deep learning architectures have often been used as a means of transparency and, as such, to shed light on the inner workings of the architectures. 
Recently, there has been a growing interest in whether or not this assumption is correct. 
In this paper we investigate the interpretability of multi-head attention in abstractive summarization, a sequence-to-sequence task for which attention does not have an intuitive alignment role, such as in machine translation. 
We first introduce three metrics to 
gain insight in the focus of attention heads and observe that these heads specialize towards relative positions, specific part-of-speech tags, and named entities. 
However, we also find that ablating and pruning these heads does not lead to a significant drop in performance,  indicating redundancy. 
By replacing the softmax activation functions with sparsemax activation functions, we find that attention heads behave seemingly more transparent: we can ablate fewer heads and heads score higher on our interpretability metrics. However, if we apply pruning to the sparsemax model we find that we can prune even more heads, raising the question whether enforced sparsity actually improves transparency. 
Finally, we find that relative positions heads seem integral to summarization performance and persistently remain after pruning.

\end{abstract}

%% !TEX root = ./main.tex
\section{Introduction}
\label{sec:introduction}

As learning algorithms become more powerful, their role in important decision making grows. 
At the same time the complexity of these learning algorithms increases \cite{schmidhuber2015deep}. 
This has given rise to a strong demand for more transparency in deep learning models, both from the general public \cite{voigt2017eu} and the research community~\cite[e.g.,][]{doshi2017towards,miller2018explanation}. 
Attention~\cite{bahdanau2014neural, luong2015effective} has gained popularity as a means of obtaining insight in the inner workings of deep neural networks~\cite[e.g.,][]{lei2017interpretable, choi2016retain,gilpin2018explaining,ghaeini2018interpreting}. 
Often examples of the attention heat map are provided to point out what attention focuses on. 
However, these examples are typically cherry-picked and leave it unclear to what extent attention can be used for transparency. 
In fact, a growing body of research has shown that one should use caution when using attention as a means for transparency. 
The majority of this work has focused on Machine Translation~\cite[e.g.,][]{vashishth2019attention} and classification~\cite[e.g.,][]{jain2019attention}, however, interpretability of attention for other tasks has not been researched as thoroughly.

We argue that in order to get a full understanding of the interpretability of attention, we should broaden our focus to other areas.  Therefore, in this work we focus on abstractive summarization, 
a task that is particularly interesting for analyzing transparency since the correspondence between input and output is less clear than in machine translation. Yet due to the sequence-to-sequence nature of the task the benefit of attention is more apparent than for language classification tasks. We specifically focus on the state-of-the-art transformer architectures~\cite{vaswani2017attention} that are commonly used for this task. By doing so, we contribute as follows: 
%\jb{Should we introduce sparsemax and pruning here? We do something different than \cite{correia2019adaptively} by enforcing the sparsest version}
\begin{enumerate}[leftmargin=*,label=(\textbf{C\arabic*}), nosep]
\item We introduce new metrics that can be used to evaluate transparency in abstractive summarization.
\item We provide insights in what attention heads in state of the art transformer architectures focus on in abstractive summarization.
\item We analyze two methods (inducing sparsity and pruning) for increasing multi-head attention interpretability applied to abstractive summarization.
% \item \jb{We use a sparsemax instead of softmax activation in attention heads to increase interpretability, and prune the resulting model to examine redundancy.}
% \item \todo{We show that established methods for increasing multi-head attention interpretability are not as faithful for abstractive summarization.}
\end{enumerate}

\section{Defining Transparency, Explainability Interpretability and Faithfulness}
\label{sec:defining_transparency}
Before we analyze transparency, explainability, interpretability and faithfulness, we need clear definitions of each of these concepts. \citet{doshi2017towards} define interpretability in Machine Learning as ``\textit{the ability to explain or to present in understandable terms to a human}''. \citet{gilpin2018explaining} define an explanation to be an answer to ``why questions'' and consider it a trade-off between interpretability and completeness. Interpretability here means \textit{understandable to humans}, whereas completeness covers how well the explanation is \textit{faithful} to the actual model mechanics. Intuitively, a \textit{transparent} model is a model in which it is clear what is happening inside. However, simply providing all parameters along with the optimization procedure violates this intuition and appears to be cheating. A transparent model should thus also be interpretable to some degree. The exact difference between explainability and transparency remains illusive. We argue that transparency addresses the \textbf{what} question: what is happening within the model? In contrast, following \citeauthor{gilpin2018explaining}'s definition, explainability addresses the \textbf{why} question: why is this output produced? Both transparency and explainability should be evaluated in terms of interpretability (how easily can we understand this explanation?) as well as faithfulness (how well does this explanation describe the system in an accurate way?).

\section{Related Work}
\label{sec:related_work}

%\mth{In this section...}

\subsection{Abstractive summarization}
The field of summarization can be divided in \textit{extractive} and \textit{abstractive} summarization. 
In this work we focus on the latter. 
The task of abstractive summarization is to construct summaries by generating new words and sentences, as opposed to directly extracting parts from the source text to add to the summary (which is extractive summarization). 
Deep learning has helped to advance abstractive summarization~\cite[e.g.,][]{rush2015neural, nallapati2016abstractive, see2017get, narayan2018don}. %
%\citet{nallapati2016abstractive} are the first to introduce a sequence-to-sequence structure for the task. After that, many have followed~\cite{see2017get} \mth{cite more}. 
With the introduction of the transformer~\cite{vaswani2017attention} and representation models like BERT~\cite{devlin2018bert}, performance on the abstractive summarization task has increased  again~\cite[e.g.,][]{gehrmann2018bottom, liu2019text}. 
We follow the state of the art and focus on transformer architectures for abstractive summarization.

\subsection{Attention as an interpretability metric}
Recently there has been a lot of interest in whether or not attention can be used to interpret or explain a model's inner functionality. Some of this work argues it can~\cite[e.g.,][]{vig2019analyzing, clark2019does, correia2019adaptively}, whereas other work argues it cannot, or one should at least be cautious~\cite[e.g.,][]{jain2019attention, serrano2019attention}. \citet{vashishth2019attention} analyze the problem over a variety of NLP tasks and conclude that it depends on the task and the importance of attention for this task. For tasks where attention does not seem to play an important role (such as text classification), attention cannot be used as interpretability metric, whereas for other tasks where attention does play a major role (such as machine translation) it can. None of these previous works focus on abstractive summarization -- a sequence-to-sequence task where attention is beneficial, yet expected to behave differently than the attention in machine translation as the correspondence between input and output is less straight forward. We close this gap in this work.

\subsection{Ablation and pruning of attention heads}
\citet{michel2019sixteen} show that a large number of attention heads can be removed without a significant drop in model performance when applying the transforme to machine translation and BERT~\cite{devlin2018bert} to natural language inference tasks. 
\citet{voita2019analyzing} introduce a pruning method that we also use in our research, hence we describe it in more detail here. 
\citeauthor{voita2019analyzing}\ apply a strategy that allows the model to retrain itself. 
They augment multi-head attention (MHA) with gates and consider them head-specific model parameters in the closed $[0,1]$ interval. 
The objective is to encourage the model to shut down heads by pushing their gates to exactly zero.
\citeauthor{voita2019analyzing}\ use a stochastic relaxation of the $L_0$ norm as follows:
\begin{align}
    \label{eq:lc_term1}
    L_C(\bm{\phi}) &= \sum_{i=1}^h (1-Q_{\bar{s}_i}(g_i = 0| \phi_i))
    \\
    \label{eq:lc_term2}
    &= \sum_{i=1}^h sigmoid(\log \alpha_j - \beta \log \frac{\epsilon}{1+\epsilon})
\end{align}
$L_C$ approximates the number of non-zero gates using the probability of these gates being non-zero. 
During training, gates are individually sampled from a Hard Concrete distribution \citep{louizos2017learning}, of which the distribution parameter $\log \alpha$ is learned. Gates are resampled for each batch. The coefficient $\lambda$ controls the weight of the regularization penalty. During inference, fixed gate values are obtained through:
\begin{align}
    \label{eq:gate_inference}
    \begin{split}
    \bm{\hat{g}} = {} &\min(\bm{1},\max(\bm{0}, \\
    &sigmoid(\log \bm{\alpha})(1+2\epsilon)-\epsilon)).
    % \bm{\theta^*} &= \bm{\theta} \odot \bm{\hat{z}}    
    \end{split}
\end{align}
\citeauthor{voita2019analyzing}\ find that the majority of the heads can be pruned with a minimal effect on overall translation performance (BLUE). 
This method has not been applied to the task of abstractive summarization.

\subsection{Sparsity in attention}
Sparsity has been used to improve the interpretability of single-head attention architectures~\cite{malaviya-etal-2018-sparse, deng2018latent, niculae2018sparsemap}. 
Commonly, these methods are based on, or extend, a sparsemax transformation \cite{martins2016softmax}.
Recently, \citet{correia2019adaptively} apply an extension of sparsemax to multi-head attention architectures for NMT. \citet{correia2019adaptively} replace the softmax in the attention heads with an $\alpha$-entmax; the higher the value for $\alpha$ the sparser. 
They show for a number of metrics that the model becomes more interpretable.

\section{Experimental Setup}
\label{sec:experimental_setup}

We use the \textit{CNN/Daily Mail} \citep{hermann2015teaching,nallapati2016abstractive} data set which consists of news articles: 287,226 training, 13,368 validation and 11,490 test pairs. Articles consist on average of 781 tokens and summaries of 3.75 sentences or 56 tokens. Following \citet{see2017get} we truncate articles to 400 words. We recover the original capitalized articles to better identify part-of-speech (POS) and named entity (NE) tags, as current state-of-the-art taggers are sensitive to capitalization.
To obtain the named entity and part-of-speech tags used for our analysis, we use out-of-the-box taggers by \citet{akbik2018coling}.\footnote{\url{https://github.com/zalandoresearch/flair}}

We adopt OpenNMT's implementation \citep{2017opennmt} of the CopyTransformer~\citep{gehrmann2018bottom} with its default hyper-paramameters. The encoder and decoder have four layers with eight heads per layer. We use two architecturally identical models. The first is an out-of-the-box model that has been pre-trained by \citet{2017opennmt}.\footnote{\url{http://opennmt.net/OpenNMT-py/Summarization.html}} The second is an identical model with different parameter initialization to investigate whether stochasticity affects the way attention heads specialize.
Table~\ref{table:baselines} shows the summarization performance of both models measured in ROUGE~\cite{lin2004rouge}.

\begin{table}[t]
\centering
\small
\begin{tabular}{llll}
\toprule
         & R-1 F1 & R-2 F1 & R-L F1 \\ \midrule
Model 1  & 38.76     & 17.13     & 36.00     \\
Model 2  & 38.81     & 16.77     & 36.28     \\
\bottomrule
\end{tabular}
\caption{ROUGE scores for two identical models with a different parameter initialization seed.}
\label{table:baselines}
\end{table}
\section{Analyzing Multi-head Attention}
\label{sec:analysis}

\begin{figure}[ht]
  \centering
  \includegraphics[width=0.45\textwidth]{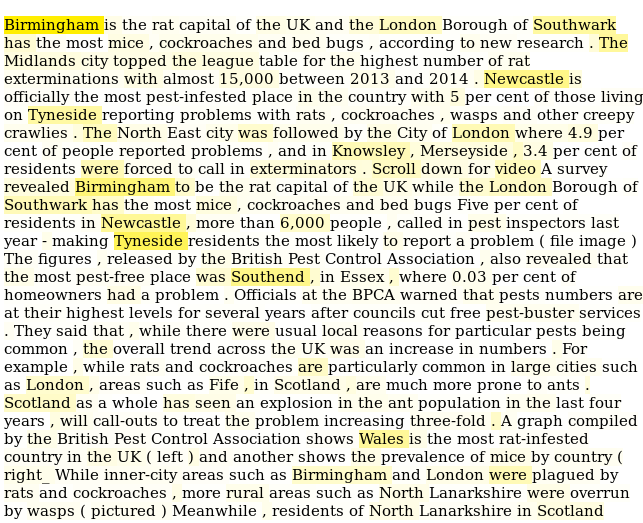}
  \caption{Attention heatmap for a decoder head that focuses on locations.}
  \label{fig:example_heatmap}
\end{figure}

We start our investigation by visually inspecting heatmaps of attention distributions to gain an intuition of its behavior (see an example heatmap in Figure \ref{fig:example_heatmap}). We observe that some heads focus on locations, people or key words---all word types that seem important to summarization. 

To quantitatively verify this observation, we design three metrics, measuring syntactic (\S\ref{sec:syntactic}), semantic (\S\ref{sec:semantic}), and positional (\S\ref{section:relative_position}) patterns, respectively. We apply these metrics to the attention distributions generated during summarization. We analyze heads from the encoder (self-attention) as well as the decoder (cross-attention) using 1K randomly sampled news articles that we pre-tag with part-of-speech and named entity tags. We exclude decoder self-attention because
\begin{enumerate*}[label=(\roman*)]
    \item its attention spans increase step-wise, causing a quantitative analysis to be significantly more complex, and
    \item encoder self-attention and cross-attention are more commonly used for interpreting MHA \cite{raganato2018analysis, clark2019does, vaswani2017attention}. 
\end{enumerate*}
We examine two identical models with a different random seed in Section \ref{sec:init} and the importance of individual heads in Section \ref{sec:analysis_ablation}.

The work presented in this section expands upon previous %\jb{unpublished} 
exploratory work by \citet{baan2019transformer} on analyzing transformer heads in abstractive summarization. Two key additions are the visualizations in Figure~\ref{fig:baseline_activations} and the ablation study in Section~\ref{sec:analysis_ablation}.

\subsection{Syntactic patterns}
\label{sec:syntactic}
\begin{figure}[h]
    \centering
    \includegraphics[width=\linewidth]{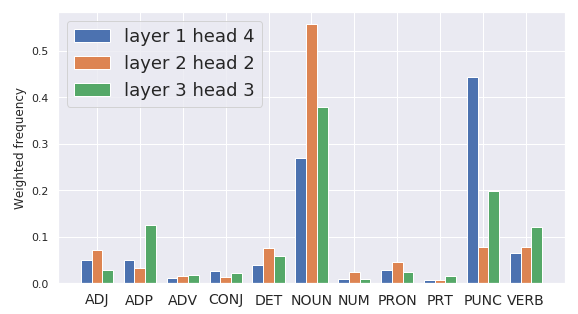}
    \caption{Attention distribution over POS tags for the top three specialized syntactic decoder heads. Large peaks appear at nouns and punctuation.}
    \label{fig:pos_distr}
\end{figure}

To quantify syntactic patterns, we compute the average KL-divergence between normalized POS tag histograms of articles and normalized attention-weighted POS tag histograms (POS-KL). 
A head focusing on specific POS tags will obtain a high POS-KL (see Figure~\ref{fig:baseline_poskl}). 
Decoder heads seem more specialized towards syntax, reflected by more heads with a high POS-KLs. 
These heads also correspond to `syntax' heads in our qualitative analysis. 
The peaks at the nouns and punctuation tags in Figure \ref{fig:pos_distr} provide insight into what exactly these heads specialize towards. 
However, the KL-divergences are relatively low (below 0.5) and we observe that there is still a considerable portion of probability mass on other syntactic categories. 
This means that heads do focus on syntax, but that it not generalize perfectly over 1000 articles. 

\subsection{Semantic patterns}
\label{sec:semantic}
We measure the ratio of attention mass on named entities (NE) to quantify semantic specialization (Figure \ref{fig:baseline_nep}). 
The KL-divergence between document and attention NE-distributions is not useful because, unlike POS tags, not every token has a NE tag. 
We find that for some attention heads, this ratio on average is more than thrice the ratio of named entity tokens in articles (0.3 and 0.1, respectively). 
The decoder head with the highest ratio lines up with the `location head' we found in our visualizations. However, even though `semantic' heads specialize, they still place large amounts of attention on other tokens. 
This appears to be due to the softmax that guarantees a smooth attention distribution. 
This is not necessarily desirable for interpretability purposes and makes reasoning about the roles of attention heads difficult. 

\subsection{Positional patterns}
\label{section:relative_position}
\begin{figure}
  \centering
  \begin{subfigure}[b]{0.49\linewidth}
  \includegraphics[width=\linewidth]{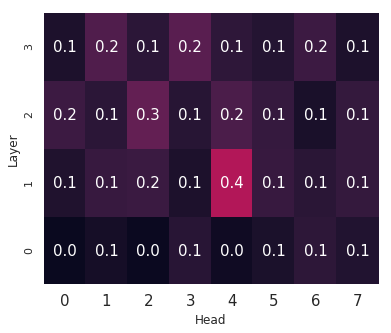}
  \caption{POS-KL (Syntax)}
  \label{fig:baseline_poskl}
  \end{subfigure}
  \begin{subfigure}[b]{0.49\linewidth}
  \includegraphics[width=\linewidth]{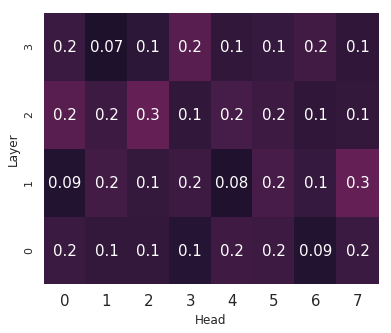}
  \caption{NE-Ratio (Semantic)}
  \label{fig:baseline_nep}
  \end{subfigure}
  \begin{subfigure}[b]{0.49\linewidth}
  \includegraphics[width=\linewidth]{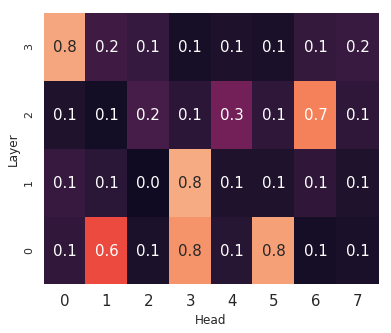}
  \caption{Relative Position}
  \label{fig:baseline_rel_pos}
  \end{subfigure}
  \begin{subfigure}[b]{0.49\linewidth}
  \includegraphics[width=\linewidth]{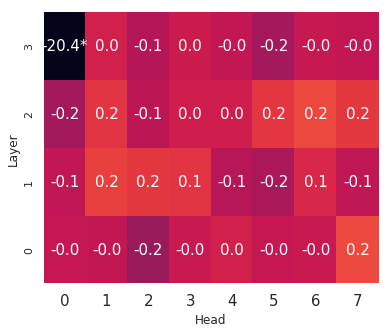}
  \caption{Ablation}
  \label{fig:baseline_ablation}
  \end{subfigure}
  \caption{Metric activations for baseline decoder heads. An asterisk in (d) depicts statistical significance for the model with ablated head with respect to the non-ablated model (t-test on 1,000 summaries).}
  \label{fig:baseline_activations}
  \vspace{-\baselineskip}
\end{figure}

We measure the ratio of each head's maximum attention weight per time step assigned to neighboring tokens (NE-ratio) and find six heads that consistently focus on preceding or succeeding tokens with a ratio of 0.8 or higher (Figure \ref{fig:baseline_rel_pos}). 
We find at least five decoder heads that focus on preceding, succeeding or currently generated tokens with a ratio of 0.7 or higher, even though there is no explicit supervisory `copy mechanism' signal. 
This behavior brings to mind the inductive bias in (Bi-)RNN architectures where tokens are explicitly processed sequentially. 
The transformer, which does not have such inductive bias and solely uses attention,  learns a similar way of processing. 
The positional activations are much higher compared to syntactic and semantic activations. This could imply that a focus on relative position is the most important specialization for abstractive summarization, or simply that its an easier task for attention heads to learn.

\subsection{Does initialization affect specialization?}
\label{sec:init}
We train two identical models with different random seeds. We find similarities as well as slight differences in specialization. 
Both models learn a similar number of relative position heads, but the second model does not contain a head that focuses on locations. This is interesting because this was exactly a head that we deemed important to the summarization process in Section \ref{sec:analysis}. 

Perhaps one model outperforms the other in terms of correctly including locations in its predicted summaries, but is worse in terms of grammar. 
We measure the per-document ROUGE scores of both models on 1K articles and compute the differences between them. 
We find an average difference of eight ROUGE points. 
We hypothesize that this at least in part explains the difference in syntactic and semantic specialization between models: both models have different strengths and weaknesses. 
These findings show that we should be careful with interpreting attention heads -- even if we could conclude that a particular instance of a trained model focuses on certain interpretable metrics, this is no guarantee that the model will always focus on this.
%The implications of this observation for transparency \jb{models are actually different? or head specializations are simply quite arbitrary?}

\subsection{Ablating heads}
\label{sec:analysis_ablation}
To further investigate the importance of attention heads and their actual impact on the resulting summaries, we perform an ablation study. 
Following \citet{michel2019sixteen}, we add a binary gate to each attention head that allows us to exclude information flow from individual attention heads. 
Interestingly, we find that not a single ablated attention head causes a statistically significant different ROUGE-1 score. \footnote{Except for the copy-head, which is jointly trained to copy tokens directly from article to summary.}
The difference in ROUGE-1 after ablating an individual head is shown in Figure~\ref{fig:baseline_ablation}. 
Significance is depicted with an asterisk. 
This demonstrates that attention heads, even those that seem to perform an interpretable task, do not affect model performance. 
It is a strong indicator that one should be careful in using the attention mechanism for transparency in abstractive summarization. 
\if0
% Not sure we need this "glue".
In the next section, we explore methods from the field of NMT to further evaluate and attempt to improve transparency in MHA.
\fi
%% !TEX root = ./main.tex

\section{Improving Transparency of MHA}
% Multi-head attention is designed to learn different representational subspaces. 
So far, we have found that it is unclear what MHA focuses on, for three reasons:
\begin{enumerate*}[label=(\roman*)]
    \item multiple heads focus on similar patterns,
    \item specialized heads still assign a considerable portion of their attention mass to tokens outside their specialization, and
    \item individual heads can be shut down without affecting the model performance.
\end{enumerate*}

% \mth{We now consider two methods that were recently proposed to improve MHA in terms of transparency: Pruning (\S\ref{sec:pruning}) and adaptive sparsity (\S\ref{sec:sparse-attention}), to our task of abstractive summarization. We evaluate the resulting models using the same specialization metrics and ablation study introduced in Section~\ref{sec:analysis}.}
%
We consider a method that was recently proposed to improve MHA in terms of interpretability, adaptive sparsity (Section~\ref{sec:sparse-attention}), for our task of abstractive summarization. 
We then apply pruning to evaluate the resulting model in terms of redundancy, as well as our specialization metrics and ablation study introduced in Section~\ref{sec:analysis}. 
We start by applying pruning to our baseline model to investigate the effects on abstractive summarization.

\subsection{Pruning attention heads}
\label{sec:pruning}
To further investigate redundancy in MHA we encourage the model to freely prune attention heads. 
We adopt a pruning strategy proposed by \citet{voita2019analyzing}. Our motivation for using this method is twofold:
First, we want to know if we observe a similar number of redundant heads for the task of abstractive summarization compared to NMT, and how removing these heads affects specialization. This is interesting since attention was designed for NMT with an intuitive meaning: alignment. 
In abstractive summarization, however, the meaning of attention is less obvious. 
Second, we expect pruning to provide additional insights into the importance of attention heads.

To remain consistent with our previous analysis, we prune encoder self-attention and decoder cross-attention heads.
Following \citet{voita2019analyzing}, we add a gate to each attention head and consider it a trainable parameter, unlike the binary gates used for ablation. 
We then encourage the model to set these gates to exactly 0 with a stochastic relaxation of the $L_0$ regularization penalty from Eq (\ref{eq:lc_term2})) to the summarization loss:
\begin{align}
    \label{eq:prune_loss}
    \text{loss} &=  \frac{1}{T}\sum_{t=1}^T -\log P_v(w^*_t) + \lambda L_C(\phi).
\end{align}
The first term is the cross-entropy, $P_v$ is the predicted distribution over the extended vocabulary, $T$ is the number of decoding time steps (predicted tokens), and $w_t^*$ is the target token for time step $t$. We test several $\lambda$'s to find an optimal value that prunes the largest amount of heads without decreasing performance.

\subsubsection{Results}
We are able to prune 34 out of 64 attention heads without a large performance impact (Table~\ref{table:pruning}). 
For some values of $\lambda$, the pruned model actually outperforms the baseline models (Table~\ref{table:baselines}). 
We believe this to be caused by the regularizing nature of the $L_C$ norm, which reduces the number of parameters and causes stronger generalization.
This confirms the hypothesis that many heads are redundant, and is in line with results from our ablation study as well as results from \citet{voita2019analyzing}. 
It is another strong indicator that attention heads should not be  used for transparency.

\begin{table}
\centering
\small
\begin{tabular}{@{\,}c@{\,\,\,}c@{\,\,\,}c@{\,\,\,}c@{\,\,\,}c@{\,}}
\toprule
$\lambda$ & \#Pruned (enc/dec) & R-1 F1 & R-2 F1 & R-L F1 \\ \midrule
0       &            & 38.76     & 17.13     & 36.00     \\
1       & 2/0        & 39.12     & 17.15     & 36.24     \\
3       & 20/14      & 38.67     & 16.66     & 36.06     \\ \bottomrule
\end{tabular}
\caption{ROUGE scores after pruning. \textit{\#0-G} shows the number of exactly-zero gates for encoder self-attention and decoder cross-attention, respectively.}
\label{table:pruning}
\end{table}

We observe that all relative position heads have been retained, comprising almost half of the remaining heads. This is interesting; these heads are interpretable, but they do not have semantic meaning for the task of summarization. Syntactic pattern activations, measured with POS-KL, are lower compared to the baseline. Semantic pattern activations, measured with NE-Ratio, are of similar magnitude. Interestingly, the interpretable `location' head was also pruned. 
% This could be explained by a phenomena that \citet{voita2019analyzing} discovers: remaining heads take on multiple specializations. However, we would expect heads to score well on both semantic as well as the syntactic metric in that case.

%\jb{Remove? 
When ablating individual heads on the pruned model, 
we observe six heads that cause a statistically significant drop in ROUGE. % when individually ablated. 
This indicates that the remaining heads are more important to model performance compared to heads in the baseline models (Section~\ref{fig:baseline_ablation}). %}
We also observe that these heads mostly correspond to strong relative position heads.
It thus appears that relative position is in fact the most important specialization to summarization performance. 
These findings are another piece of evidence for redundancy. 
More importantly, even though heads appear to be interpretable (such as the `location' head), they do not necessarily contribute to the overall summarization performance. 
Thus, one should be careful with using them for transparency. 
In the next section we investigate a method that was recently introduced to increase the interpretability of multi-head attention in NMT.

\subsection{Sparse attention}
\label{sec:sparse-attention}
\citet{correia2019adaptively} propose to replace the softmax activation function with $\alpha$-entmax to improve the interpretability of multi-head attention. In an attempt to improve the interpretability even further, we use the sparsest case of $\alpha$-entmax instead: the sparsemax transformation. 

In Section \ref{sec:analysis} we observed that specialized heads place a considerable amount of attention mass on non-related tokens. The sparsemax activation function seems well suited to address this problem. Sparsemax projects an input vector $\textbf{z}$ onto the probability simplex and is formally defined as:
\begin{align}
    \label{eq:sparsemax}
    \mbox{}\hspace*{-2mm}
    \text{sparsemax}(\textbf{z}) &= \underset{\textbf{p} \in \Delta^{K-1}}{\text{argmin}} ||\textbf{p}-\textbf{z} ||^2 \\
    \label{eq:simplex}
    \Delta^{K-1} &= \{\mathbf{p} \in \mathbb{R}^K \mid \textbf{1}^T\ \mathbf{p}=1, \geq \textbf{0}\},
\end{align}
resulting in the following attention function:
\begin{align}
    \text{Attn}(Q,K,V) &= \text{sparsemax}\left(\frac{QK^T}{\sqrt{d_k}}\right)V.
\end{align}
We first apply this modification to the \textbf{Enc}oder (\textit{Sparse-Enc} model). We then extend it to the entire model. However, we observe that we need to exclude the \textbf{T}op \textbf{L}ayer of the decoder (\textit{Sparse-TL} model) to prevent performance from collapsing. We hypothesize that this is due to interference with the copy-head, located in the top layer. To further test this hypothesis, we individually exclude the \textbf{C}opy \textbf{H}ead (\textit{Sparse-CH} model) instead of the entire top layer. 

\subsubsection{Results}
We can push 97\% of all attention weights to zero for all sparsemax heads in the encoder without affecting performance. Sparse-TL and Sparse-CH both perform competitively with a minor drop in ROUGE (Table \ref{table:sparsemax}). This is in line with findings in NMT by \citet{correia2019adaptively}, although their $\alpha$-entmax model is free to choose the degree of sparsity, unlike ours. We find that the copy head is indeed the cause of performance collapse as applying sparsemax results in roughly half the ROUGE score. We hypothesize that the sparsemax activation interferes with the loss computation or OOV token sampling of the copy head, but want to investigate this in more detail. 

\begin{table}[t]
\centering
\small
\begin{tabular}{lccc}
\toprule
           & R-1 F1    & R-2 F1   & R-L F1    \\ \midrule
Sparse-Enc & 38.73     & 16.68     & 35.64      \\
Sparse-TL  & 38.51     & 16.87     & 35.66      \\
Sparse-CH  & 38.30     & 16.47     & 35.35      \\ \bottomrule
\end{tabular}
\caption{ROUGE scores for using sparsemax on all \textbf{Enc}oder heads, all heads except for the decoder \textbf{T}op \textbf{L}ayer, and the entire model except for the \textbf{C}opy \textbf{H}ead. }
\label{table:sparsemax}
\end{table}

\begin{figure}[t]
  \centering
  \begin{subfigure}[b]{0.49\linewidth}
  \includegraphics[width=\linewidth]{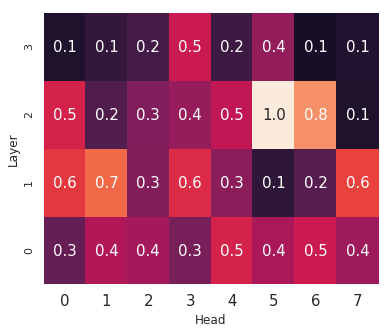}
  \caption{POS-KL (Syntax)}
  \label{fig:sparsemax_poskl}
  \end{subfigure}
  \begin{subfigure}[b]{0.49\linewidth}
  \includegraphics[width=\linewidth]{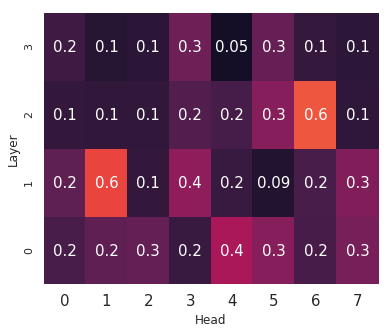}
  \caption{NE-Ratio (Semantic)}
  \label{fig:sparsemax_nep}
  \end{subfigure}
  \quad
  \begin{subfigure}[b]{0.49\linewidth}
  \includegraphics[width=\linewidth]{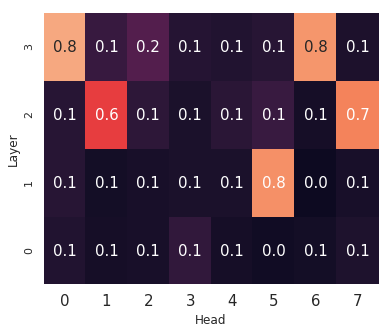}
  \caption{Relative Position}
  \label{fig:sparsemax_rel_pos}
  \end{subfigure}
  \begin{subfigure}[b]{0.49\linewidth}
  \includegraphics[width=\linewidth]{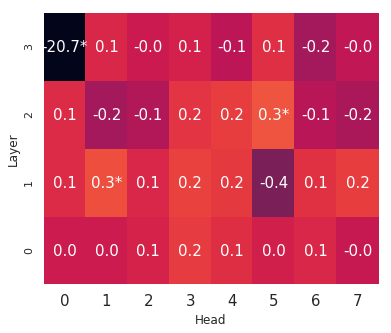}
  \caption{Ablation}
  \label{fig:sparsemax_ablation}
  \end{subfigure}
  \caption{Metric activation for Sparse-TL decoder heads.}
  \label{fig:sparsemax_activations}
  \vspace{-\baselineskip}
\end{figure}

\begin{figure}[t]
  \centering
  \includegraphics[width=0.45\textwidth]{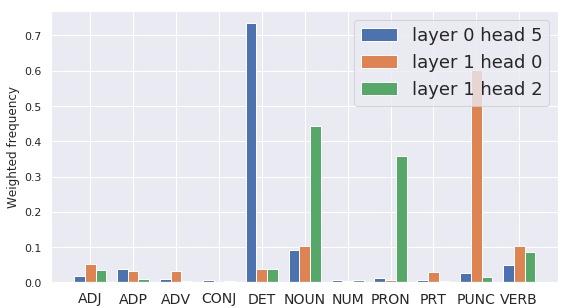}
  \caption{Highly focused attention distributions over POS tags. We show the top three specialized syntactic encoder heads in the Sparse-TL model.}
  \label{fig:sparsemax_pos_distr}
\end{figure}

Figure \ref{fig:sparsemax_activations} shows stronger activation on syntactic (POS-KL) and semantic (NE-Ratio) patterns compared to the baseline (Figure \ref{fig:baseline_activations}). 
Upon closer inspection of heads with a high syntactic activation
we see significantly more peaked distributions over POS tags (Figure \ref{fig:sparsemax_pos_distr} shows three encoder heads with the highest POS-KL). Interestingly, an unseen strong syntactic pattern emerges that focuses on determinants or pronouns. This again stresses that different specializations can emerge in different models.

Do stronger activations on our metrics imply that individual heads have become more important to summarization? We observe nine heads that statistically significantly affect performance when ablated. It thus seems that not only sparsemax heads specialize more distinctly, but that the impact of individual heads has also increased.

\subsection{Pruning sparse attention heads}
We have discovered that we can prune roughly half the attention heads and that a sparsemax activation function appears to improve transparency.
%of MHA through more distinct specialization and more heads that statistically significantly impact performance.
If we now prune a transformer model with sparse attention, we would expect fewer heads that can be pruned. After all, the sparsemax-activated heads appear more interpretable as well as faithful.

We prune the Sparse-TL model of which the top decoder layer heads retain their softmax because \begin{enumerate*}[label=(\roman*)]
    \item its performance is superior, and
    \item we want to investigate whether there is a preference for pruning heads with either a softmax or a sparsemax activation function. 
\end{enumerate*}

\subsubsection{Results}
Surprisingly, we are able to prune an even larger amount of heads in our sparsemax model compared to the baseline model: 43 out of 64 heads (Table \ref{table:pruningsparsemax}). All heads in the decoder top layer are retained. Perhaps the freedom that a softmax activation provides by allowing for more diffuse distributions is important, or the heads in the top decoder layer were incidentally more important.

\begin{table}[h]
\centering
\small
\begin{tabular}{@{\,}c@{\,\,\,}c@{\,\,\,}c@{\,\,\,}c@{\,\,\,}c@{\,}}
\toprule
\#Pruned (enc/dec)    & R-1 F1 & R-2 F1 & R-L F1  \\ \midrule
22/21      & 38.45     & 16.63     & 35.39      \\ \bottomrule
\end{tabular}
\caption{Results for a sparse transformer (Sparse-TL) after pruning on ROUGE. $\lambda$ is emperically set to 2.}
\label{table:pruningsparsemax}
\end{table}

\begin{figure}[t]
  \centering
  \begin{subfigure}[b]{0.49\linewidth}
  \includegraphics[width=\linewidth]{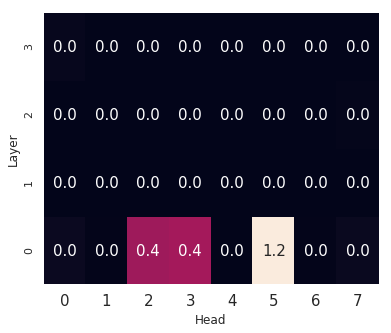}
  \caption{POS-KL (Syntax)}
  \label{fig:pruned_sparsemax_poskl_enc}
  \end{subfigure}
  \begin{subfigure}[b]{0.49\linewidth}
  \includegraphics[width=\linewidth]{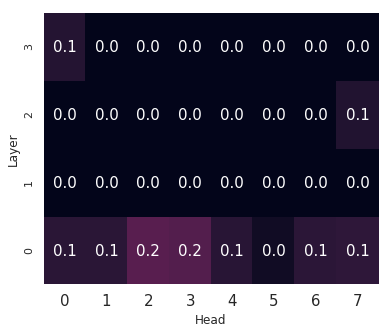}
  \caption{NE-Ratio (Semantic)}
  \label{fig:pruned_sparsemax_nep_enc}
  \end{subfigure}
  \quad
  \begin{subfigure}[b]{0.49\linewidth}
  \includegraphics[width=\linewidth]{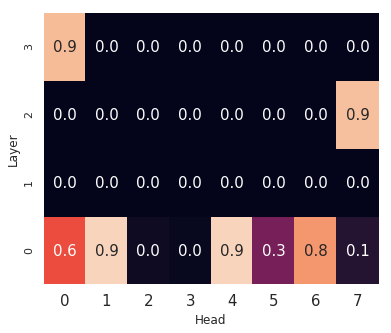}
  \caption{Relative Position}
  \label{fig:pruned_sparsemax_rel_pos_enc}
  \end{subfigure}
  \begin{subfigure}[b]{0.49\linewidth}
  \includegraphics[width=\linewidth]{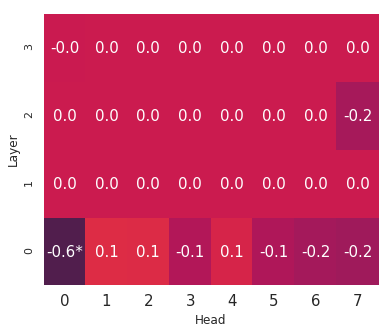}
  \caption{Ablation}
  \label{fig:pruned_sparsemax_ablation_enc}
  \end{subfigure}
  \caption{Metric activation for Sparse-TL encoder heads after pruning.}
  \label{fig:pruned_sparsemax_enc}
  \vspace{-\baselineskip}
\end{figure}

We find that five out of the ten remaining encoder heads focus strongly on relative positions (Figure \ref{fig:pruned_sparsemax_enc}). 
The remaining encoder heads score high on syntactic patterns. However, upon inspection of the strongest syntactic encoder head we discover it that almost exclusively focuses on the word `the'. 
This does not appear to be a relevant specialization to summarization, but nonetheless this head is one of the few encoder heads that survived the pruning process. 

\begin{figure}[t]
  \centering
  \begin{subfigure}[b]{0.49\linewidth}
  \includegraphics[width=\linewidth]{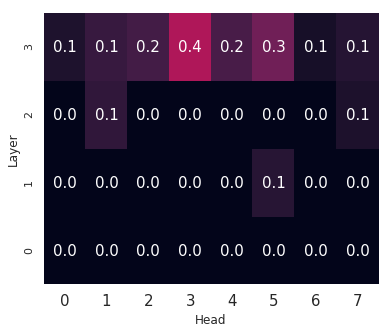}
  \caption{POS-KL (Syntax)}
  \label{fig:pruned_sparsemax_poskl}
  \end{subfigure}
  \begin{subfigure}[b]{0.49\linewidth}
  \includegraphics[width=\linewidth]{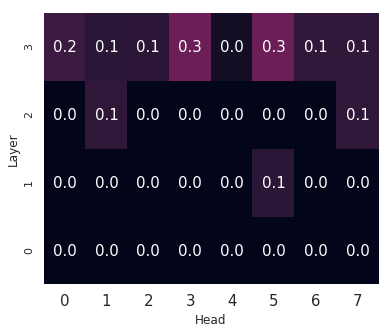}
  \caption{NE-Ratio (Semantic)}
  \label{fig:pruned_sparsemax_nep}
  \end{subfigure}
  \quad
  \begin{subfigure}[b]{0.49\linewidth}
  \includegraphics[width=\linewidth]{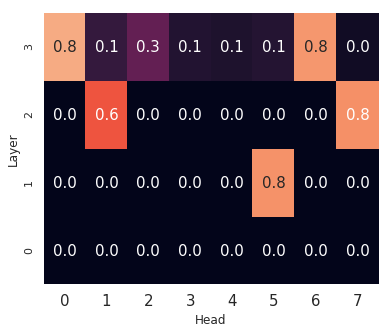}
  \caption{Relative Position}
  \label{fig:pruned_sparsemax_rel_pos}
  \end{subfigure}
  \begin{subfigure}[b]{0.49\linewidth}
  \includegraphics[width=\linewidth]{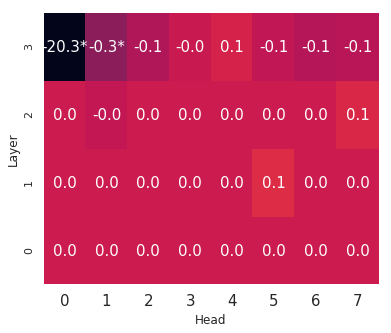}
  \caption{Ablation}
  \label{fig:pruned_sparsemax_ablation}
  \end{subfigure}
  \caption{Metric activation for Sparse-TL decoder heads after pruning.}
  \label{fig:pruned_sparsemax}
  \vspace{-\baselineskip}
\end{figure}

Similar to the encoder, the majority of decoder heads focus on relative position (Figure \ref{fig:pruned_sparsemax}). The others respond slightly higher to semantic as well as syntactic metrics compared to the baseline, but not by a large margin. For heads in the top layer this could be expected, as they use the same softmax activation function as heads in the baseline.

% However, only two out of the 21 retained heads produce a statistically significant difference when ablated. Both heads are strong relative position heads. This low number is surprising: judging from ablations for pruning as well as sparsemax models we would have expected most retained heads to cause a statistically significant difference. 
% We bring forward two possible explanations for this. The first is straightforward: a higher value for $\lambda$ might result in more heads being pruned without affecting performance. The second is that forcing both attention heads as well as attention weights to zero has somehow pushed to model to rely less on multi-head attention. This could have made it more robust to the ablation of individual attention heads. 

We can conclude that sparsemax appears to improve the interpretability as well as faithfulness of MHA. It scores high on our specialization metrics, and contains more heads with a statistically significant impact on performance. However, when we prune a sparsemax model, we are able to prune even more heads than we could in the baseline model. Additionally, most of the semantic and syntactic specialized heads disappear. The remaining heads are predominantly relative position heads. This gives rise to an important question: does sparsity in multi-head attention actually improve transparency?

% A possible explanation for this larger number of removed heads compared to non-sparsemax models is that sparsemax forces representations to be less distributed over attention heads. This might cause multiple heads with the same distinct specialization. These can then be easily pruned.

% We can conclude that sparsemax appears to improve interpretability as well as faithfulness by scoring high on specialization metrics as well as 
% reducing the number of attention heads, and achievied higher degrees of specialization compared to the baseline. However, compared to our non pruned sparsemax model, the specialization on POS-KL and NE-ratio seems to have decreased. In terms of faithfulness, we are unable to increase the effect of all remaining individual attention heads. Is pruned sparsemax MHA transparent? We argue that it offers better transparency compared to vanilla multi-headed attention. However, it is far from completely transparent, as we still do not fully understand how the model as a whole uses attention heads to arrive at a prediction.

%% !TEX root = ./main.tex
\section{Discussion}
How do we evaluate whether attention can be used as means for transparency? This question is raised over and over again, and is very difficult to answer. This is illustrated by a large body of recent work that adresses this question (see Section \ref{sec:related_work}). Quantifying specialization in attention heads, pruning and ablation studies provide more insights, but still result in contradictory observations. 
A promising recent line of work focuses on adversarial attention attacks \cite{jain2019attention, serrano2019attention, vashishth2019attention}, but this is not yet applied to sequence-to-sequence tasks. We believe this to be a promising next step in better understanding attention as means for transparency. Finally, in this work, as well as in most related work, we assume that representations learned within a transformer model correspond to a (contextualized representation of) the input token at that position. This assumption should be properly investigated.

\section{Conclusion}
\label{sec:conclusion}
We have investigated to what extent multi-head attention in abstractive summarization is transparent. We have introduced quantitative metrics that showed that multi-head attention is partially interpretable. However, we have also shown that all individual heads can be ablated without a significant drop in performance, indicating that one should be very careful using the attention mechanism for transparency in abstractive summarization. Replacing the softmax activation function by a sparsemax activation function resulted in improved scores on our interpretability metrics, and fewer heads that could be ablated without decreasing summarization performance. However, in this setting more heads can be pruned. In all our experiments we find that relative position heads seem integral to performance and persistently remain after pruning.
%This opens the door towards more research to answer \mth{these questions}

Taking all our findings and related work into account, we believe that for multi-head attention to be transparent, it should adhere to the following criteria:
\begin{enumerate*}[label=(\roman*)]
    \item multi-head attention should have a minimum number of heads,
    \item heads should have no overlap in specialization but focus on distinct representational subspaces, and
    \item we need the right metrics to measure interpretability.
\end{enumerate*}

%We show that some attention heads specialize towards relative position, POS tags and NE tags. However, specialization does not fully generalize across articles or models. We also show that ablating individual attention heads does not statistically significantly affect summarization performance.

%We successfully prune 34 out of 64 attention heads using a stochastic relaxation of the $L_0$ norm as regularization penalty. The resulting model outperforms the baseline on ROUGE and is more interpretable due to this decrease of components. It also improves on faithfulness, as ablating several heads cause statistically significant changes.

%We replace the softmax activation over attention scores with a sparsemax activation, resulting in drastic improvement on our specialization metrics, competitive performance on ROUGE and several heads causing statistically significant changes when ablated.

% Combining the pruning strategy with a sparsemax activation function demonstrates the ability to \begin{enumerate*}
%     \item prune the largest number of heads yet,
%     \item increase specialization of remaining heads, and
%     \item perform competitively on ROUGE.
% \end{enumerate*} 

% In conclusion, we believe there are two important implications from our work.
% \begin{enumerate}
%     \item The research community should be careful with using attention heads for transparency, as there are both problems with interpretability as well as faithfulness.
%     \item Addressing redundancy in multi-headed attention can improve on both transparency and performance.
% \end{enumerate}

\bibliography{acl2019}
\bibliographystyle{acl_natbib}

\end{document}